\documentclass[runningheads]{llncs}

\usepackage{eccv}

\usepackage{eccvabbrv}

\usepackage{graphicx}
\usepackage{booktabs}
\usepackage{tcolorbox}
\usepackage{adjustbox}
\usepackage{booktabs}
\usepackage{graphicx}
\usepackage{colortbl}
\usepackage{xcolor}
\usepackage{subcaption}

\usepackage[accsupp]{axessibility}  

\usepackage{hyperref}

\begin{document}

\title{Learning from Medical Entity Trees: An Entity-Centric Medical Data Engineering Framework for MLLMs} 

\titlerunning{Learning from Medical Entity Trees}

\author{
Jianghang Lin\inst{1}\textsuperscript{*} \and
Haihua Yang\inst{2}\textsuperscript{*} \and
Deli Yu\inst{2} \and
Kai Wu\inst{2} \and
Kai Ye\inst{1} \and
Jinghao Lin\inst{2} \and
Zihan Wang\inst{3} \and
Yuhang Wu\inst{1} \and
Liujuan Cao\inst{1}\textsuperscript{\dag}
}

\authorrunning{J. Lin, H. Yang et al.}

\institute{
Key Laboratory of Multimedia Trusted Perception and Efficient Computing,
Ministry of Education of China, Xiamen University, China
\and
ByteDance
\and
Northeastern University\\
\email{\textsuperscript{*}Equal contribution. \enspace
\textsuperscript{\dag}Corresponding author.}
}

\maketitle

\begin{abstract}
Multimodal Large Language Models (MLLMs) have shown transformative potential in medical applications, yet their performance is hindered by conventional data curation strategies that rely on coarse-grained partitioning by modality or department.
Such fragmented approaches fail to capture the hierarchical and interconnected nature of clinical medical knowledge, limiting the models’ ability to perform fine-grained recognition and complex reasoning.
In this paper, we propose a novel Entity-Centric Medical Data Engineering framework.
We automatically extract entities from authoritative medical literature to construct a Medical Entity Tree (MET), a hierarchical structure that systematically encodes diseases, anatomical structures, modalities, and symptoms into a unified knowledge repository.
Building upon the MET, we propose an advanced data engine that includes: (1) node-guided retrieval to anchor raw data to specific medical concepts, (2) a two-stage hybrid filtering and alignment pipeline to ensure precise visual-semantic correspondence, and (3) knowledge-aware data synthesis to generate enriched captions and targeted reasoning VQA pairs, leveraging structural constraints.
Extensive evaluations across six medical benchmarks demonstrate that our approach significantly enhances the medical capabilities of general-purpose MLLMs, improving their ability to handle complex clinical queries and achieve state-of-the-art performance in diverse medical contexts.
\end{abstract}    
\section{Introduction}
\label{sec:intro}
\begin{figure}[!t]
\centering
\includegraphics[width=0.95\linewidth]{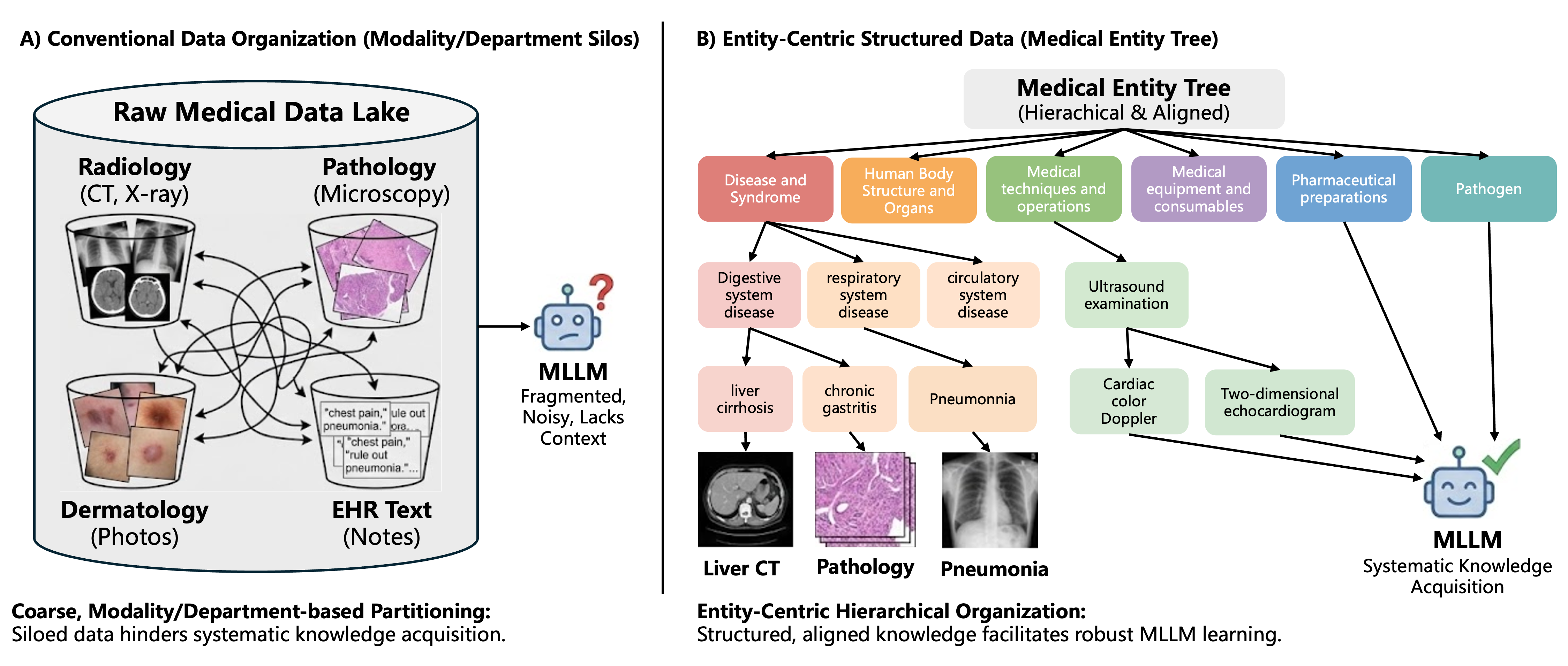}
\caption{\textbf{Paradigmatic shift from modality-based partitioning to knowledge-centered data organization.} A) Traditional data curation often treats medical knowledge as a flat collection of images and labels. B) Our framework mimics the systematic pedagogical path of human medical expertise by anchoring multimodal data to a structured Medical Entity Tree, enhancing the coverage of long-tail medical entities and fine-grained reasoning.}
\label{fig:comparison}
\vspace{-20pt}
\end{figure}
Multimodal Large Language Models (MLLMs)~\cite{hurst2024gpt,wang2024qwen2,liu2023visual,team2023gemini,shi2026medxiaohe} have shown remarkable potential in clinical applications~\cite{li2023llava,sellergren2025medgemma,zhang2023huatuogpt,xu2025lingshu,wang2026deepmed} such as assisted diagnosis, medical visual question answering, and medical report generation. However, their performance remains heavily reliant on the availability of high-quality, large-scale multimodal data. Despite the abundance of medical datasets~\cite{yue2024mmmu,lau2018dataset,zhang2023pmc,liu2021slake,he2020pathvqa,hu2024omnimedvqa}, as shown in Fig.\,\ref{fig:comparison}, current data curation efforts often rely on coarse-grained categorization, typically partitioning data by modality (e.g., X-ray, CT, MRI)~\cite{chowdhury2020can,yan2018deeplesion,lou2025sdr} or broad clinical departments (e.g., Dermatology, Ophthalmology)~\cite{tschandl2018ham10000,nakayama2024brset}. This approach raises a fundamental question: \textit{Is such fragmented and superficial data organization sufficient for MLLMs to internalize complex medical knowledge?}
Medical knowledge, in practice, is not a flat collection of images and labels. It forms an intricate web of interconnections, spanning disease pathology, anatomical structures, radiological signs, and therapeutic procedures. Coarse-grained categorization assumes that ``modality'' or ``department'' alone defines the boundaries of knowledge. However, this simplification fails to capture the multi-dimensional nature of clinical reasoning, limiting MLLMs' ability to perform fine-grained recognition, causal inference, and provide comprehensive coverage of long-tail medical entities.
In contrast, human medical expertise is developed through a structured and hierarchical learning process. For instance, when studying a condition like ``liver cirrhosis,'' a practitioner follows a systematic path—from understanding pathophysiology and anatomical changes to diagnosing modalities, typical signs, and differential diagnosis—incorporating authoritative knowledge from textbooks and clinical guidelines. This prompts a natural hypothesis: \textit{Can we empower MLLMs with a similar systematic learning paradigm?} Given that medical literature (such as textbooks and research papers) provides the most comprehensive record of human medical knowledge, there is a need for a mechanism to transform this ``textbook-level knowledge structure'' into a ``model-ready data organization.''
To address this challenge, we propose a novel \textbf{Entity-Centric Medical Data Engineering} framework. We automatically extract medical entities from open-source medical textbooks and research papers to construct a Medical Entity Tree (MET). This hierarchical structure organizes diseases, anatomical sites, modalities, symptoms, and surgical procedures at multiple levels. In the MET, each node represents more than just a terminology; it encodes a ``knowledge context'' through its hierarchical and associative relationships, transforming fragmented data into a structured knowledge repository. These entity nodes serve as unified ``query and alignment units,'' guiding subsequent data collection, cleaning, and annotation.
Building on the MET, we introduce a systematic pipeline for data acquisition and synthesis. Initially, we retrieve relevant multimodal data from web and open-source repositories, using entity nodes as anchors. We then apply rigorous filtering and re-captioning techniques to ensure precise alignment between entities and visual features. Crucially, we leverage the structural constraints of the entity tree to perform knowledge-aware data augmentation and synthesis (e.g., generating structure-constrained reasoning Q\&A pairs based on specific ``disease-sign-modality'' chains). This results in a task-oriented, knowledge-aligned, and scalable training corpus.
We evaluate our synthesized data by fine-tuning several mainstream open-source MLLM backbones. Experimental results across six authoritative medical benchmarks—MMMU-Med~\cite{yue2024mmmu}, VQA-RAD~\cite{lau2018dataset}, SLAKE~\cite{liu2021slake}, PathVQA~\cite{he2020pathvqa}, PMC-VQA~\cite{zhang2023pmc}, and OmniMedVQA~\cite{hu2024omnimedvqa}—demonstrate significant and consistent improvements in performance. Our model achieves an average accuracy of 69.16\%, surpassing state-of-the-art models such as Lingshu-7B~\cite{xu2025lingshu}. Notable improvements are seen in expert-level reasoning tasks, such as MMMU-Med (73.77\%), as well as in large-scale datasets like OmniMedVQA (83.36\%). Additionally, our model excels at recognizing both frequent and rare medical entities, showing particularly higher accuracy on sparsely represented conditions. The generalization experiments confirm consistent performance gains across different architectures, validating the effectiveness of our entity-driven data synthesis.
%
\section{Related Work}
\label{sec:related_work}
\subsection{Multimodal Large Language Models}
The development of multimodal intelligence has shifted from discriminative Vision-Language Models (VLMs) to generative Multimodal Large Language Models (MLLMs).
Early VLMs, such as CLIP~\cite{radford2021learning}, ALIGN~\cite{jia2021scaling}, and SigLIP~\cite{zhai2023sigmoid}, aligned visual and textual representations using contrastive learning for zero-shot classification.
MLLMs, however, leverage the reasoning and generative capabilities of Large Language Models (LLMs).
This shift began with modular connectors designed to compress high-dimensional visual features into a fixed set of tokens.
Notable examples include the Q-Former in BLIP-2~\cite{li2023blip}, which uses a querying transformer to extract relevant visual semantics, and the Perceiver Resampler in Flamingo~\cite{alayrac2022flamingo}, which employs cross-attention to handle variable-length inputs. Later, the projection-based approach popularized by LLaVA~\cite{liu2023visual} utilized a simple MLP to treat visual embeddings as soft prompts for LLMs. Subsequent advancements focused on scaling and resolution, with Qwen2-VL~\cite{wang2024qwen2} introducing dynamic resolution handling and InternVL~\cite{chen2024internvl} scaling the vision backbone to billions of parameters to capture intricate local details.
This architectural evolution has been driven by a shift from noisy pre-training on image-text pairs to high-quality visual instruction tuning and synthetic reasoning tasks.
In the medical field, these developments are adapted to handle clinical precision and modality-specific nuances. 
Early models like LLaVA-Med~\cite{li2023llava} and Med-PaLM~\cite{singhal2023large} applied general-purpose MLLM architectures to medical data with curated datasets like PMC-15M~\cite{zhang2023biomedclip}. 
However, challenges remain in fine-grained feature perception and reducing clinical hallucinations~\cite{ye2025multimodal}.
Lingshu~\cite{xu2025lingshu} is the current state-of-the-art, integrating over 12 medical modalities and using reinforcement learning with verifiable rewards to ensure diagnostic accuracy. 
Recent research has expanded to 3D reasoning for MRI and CT scans, emphasizing the importance of factual generation and Retrieval-Augmented Generation (RAG) to ensure clinical insights are grounded in visual evidence and medical knowledge. 
Despite these advancements, the current approach still relies on fragmented data organization, limiting data efficiency. Therefore, a unified, knowledge-centered data strategy is essential for further progress.
\subsection{Medical Datasets}
The landscape of medical datasets can be broadly categorized into expert-annotated clinical benchmarks, large-scale multimodal corpora, and the emerging paradigm of knowledge-structured data. 
Expert-annotated benchmarks, such as VQA-RAD~\cite{lau2018dataset} and SLAKE~\cite{liu2021slake}, provide high-fidelity, clinician-generated question-answer pairs that ensure natural linguistic diversity and clinical relevance. 
Similarly, VQA-Med-2019~\cite{ben2019vqa} and PathVQA~\cite{he2020pathvqa} offer structured evaluations for radiology and pathology, respectively, often categorizing questions by modality, organ system, or abnormality.
With the rise of MLLMs, large-scale multimodal corpora like PMC-VQA~\cite{zhang2023pmc}, OmniMedVQA~\cite{hu2024omnimedvqa} and Derm1M~\cite{yan2025derm1m} have scaled significantly to support pre-training, however, these datasets often rely on coarse-grained partitioning by modality or broad clinical departments.
This fragmented organization treats medical knowledge as a flat collection of images and labels, failing to capture the hierarchical and interconnected nature of clinical reasoning.
Recent research in general domain has explored incorporating external knowledge through retrieval-based or implicit methods. 
For instance, RAVQA-V2~\cite{lin2023fine} and TRIG~\cite{gao2022transform} utilize web search or Wikidata to align visual and textual information, while Prophet~\cite{shao2023prompting} and SKP~\cite{wang2024soft} leverage MLLMs as implicit knowledge bases or attention mechanisms to enhance reasoning. 
However, these methods often remain bottlenecked by the siloed nature of the underlying data.
In contrast, our approach shifts the paradigm toward a knowledge-centered organization by anchoring data to a structured Medical Entity Tree.
By transforming textbook-level hierarchical knowledge into a model-ready repository, our framework ensures comprehensive coverage of long-tail entities and facilitates fine-grained reasoning that traditional modality-based partitions lack.
\section{Methodology}
\label{sec:methodology}
\begin{figure*}[!t]
\centering
\includegraphics[width=1.0\linewidth]{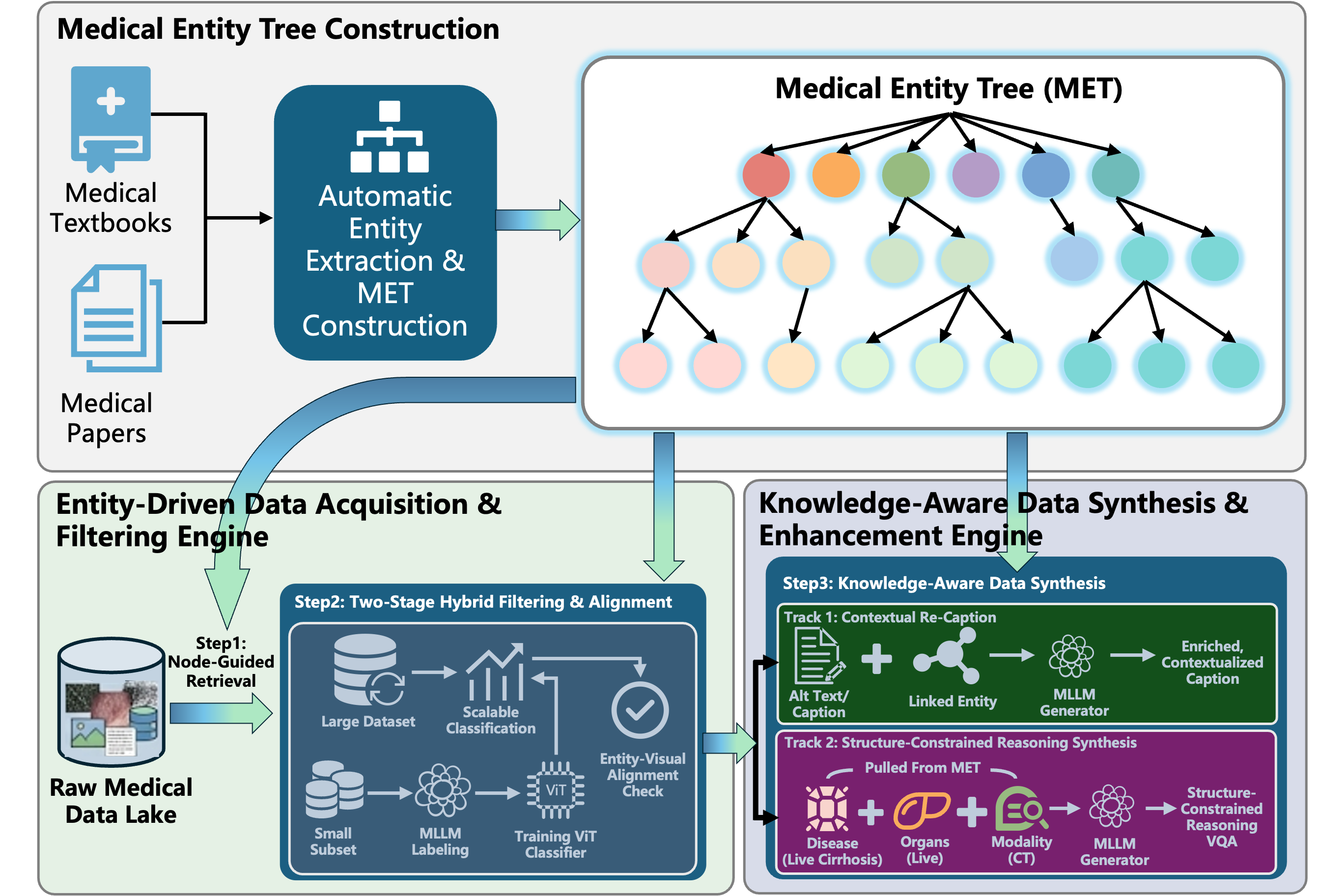}
\caption{\textbf{Overview of the proposed Entity-Centric Medical Data Engineering framework.} The pipeline consists of three core stages: (1) \textbf{Medical Entity Tree Construction}, where entities are extracted from medical textbooks and papers to build a hierarchical knowledge structure; (2) \textbf{Entity-Driven Data Acquisition \& Filtering}, featuring a two-stage hybrid alignment process (MLLM labeling and ViT-based scalable classification); and (3) \textbf{Knowledge-Aware Data Synthesis \& Enhancement}, which generates contextualized captions and structure-constrained reasoning samples to ensure high-fidelity knowledge alignment for MLLM training.}
\label{fig:framework}
\vspace{-20pt}
\end{figure*}
\subsection{Construction of Medical Entity Tree}
Our methodology operates under the premise that \textit{``open-source medical literature encapsulates the totality of currently known human medical knowledge.''} Consequently, our objective is to distill a structured ontology directly from this comprehensive repository. We developed a multi-stage pipeline evolving from coarse extraction to fine-grained refinement to mine entities from these materials, effectively navigating the trade-off between extraction quality and computational efficiency.
\subsubsection{Stage 1: High-Efficiency Entity Extraction via Prompt Engineering}
Initially, we leveraged Large Language Models (LLMs) to identify medical entity nouns from the massive corpus. To mitigate inference latency, we implemented a \textit{Batch Chunking} strategy, identifying entities within consolidated multi-sentence input units.
To ensure precision and alignment within this batched framework, we designed a rigorous prompt (detailed in \textbf{Supp. A}) that enforces strict output constraints. 
Specifically, the prompt instructs the model to: (1) filter non-medical terms exclusively; (2) eliminate duplicates; (3) explicitly output a ``None'' token for sentences lacking medical entities; and (4) strictly adhere to a JSON output format.
This design significantly facilitates downstream parsing. By enforcing structured output with explicit null-case handling, this approach yielded a processing speedup of approximately $30\times$ while maintaining high extraction recall and effectively filtering non-medical noise.
This stage provides a large number of candidate entities for the stage 3 expansion.
\subsubsection{Stage 2: Joint Typing and Hierarchical Clustering} 
To evolve from a flat noun list to a structured system, the second stage advances the prompt engineering strategy to simultaneously perform entity extraction and type abstraction.
The prompt design for this stage addresses specific challenges inherent in medical texts. As detailed in \textbf{Supp. A}, the instructions mandate the model to: (1) handle noisy characters often introduced by OCR or formatting errors; (2) ensure appropriate entity granularity by segmenting overly long phrases; and (3) generate a joint output of entity types and names in a structured key-value pair format (e.g., \texttt{<EntityType>:<EntityName>}).
Post-extraction, we enforced a rigorous cleansing protocol: entities appearing fewer than 10 times and types containing fewer than 5 entities were filtered to reduce noise. Subsequently, we applied K-Means clustering~\cite{mcqueen1967some} on entity embeddings, utilizing the Silhouette Coefficient~\cite{rousseeuw1987silhouettes} to dynamically optimize the cluster count for defining secondary types. By applying a bottom-up, frequency-weighted aggregation of embeddings, we recursively constructed a multi-level entity hierarchy.
Finally, we manually pruned and confirmed a core taxonomy.
\subsubsection{Stage 3: Controlled Expansion via Tree Attachment} 
Upon completing Stages 2, we obtained a concise, high-quality core taxonomy. Stage 3 focuses on scaling this repository by integrating entities extracted from the broader corpus.
\textit{Incremental Tree Attachment.} Instead of rebuilding the tree from scratch, we adopted an Incremental Tree Attachment strategy. We treat the core taxonomy as a fixed skeleton and utilize the semantic reasoning capabilities of LLMs to graft new candidate entities onto the most appropriate nodes. To ensure the integrity of the existing structure, the prompt (see \textbf{Supp. A}) positions the LLM as a taxonomic expert. It is strictly mandated to identify the correct insertion path without altering existing node names and must provide an explicit reasoning chain (wrapped in \texttt{<Reasoning>} tags) for every insertion or rejection to ensure interpretability and traceability.
\textit{Optimization via Deferred Insertion.} As the taxonomy expands, fitting the entire tree structure into the LLM context window becomes infeasible. We addressed this through a ``Deferred Insertion'' strategy: the high-quality core tree is frozen as a context anchor. For subsequent insertions, the LLM predicts the insertion path relative to this frozen core. These operations are buffered and executed in batch updates.
\textit{Conflict Resolution via ReAct Agent.} While deferred insertion ensures scalability, it introduces structural ambiguity: a single entity may develop high affinity with multiple parent nodes (e.g., \textit{Kluver-Bucy syndrome} appearing under both \textit{Neurological Disorders} and \textit{Digestive Symptoms}). This creates logical conflicts where the taxonomy degrades into a graph structure with overlapping definitions. Solely relying on the internal parametric knowledge of LLMs to resolve these conflicts often leads to hallucinations.
To address this, we implemented a \textit{ReAct (Reasoning + Acting) Agent} framework~\cite{yao2022react}. We designed a specialized prompt (refer to \textbf{Supp. A}) that forces the agent to search for exact medical definitions via external tools (e.g., Google or Wikipedia) and adjudicate conflicts based on two core principles: (1) \textit{Principle of Etiological Dominance:} Classifications based on pathological mechanisms or anatomical locations take precedence over clinical symptoms. (2) \textit{Principle of Specificity:} If one parent category is a subset of another and describes the entity more accurately, the more specific category is preferred.
The transition from a passive LLM to an active Agent offers three critical benefits: (1) \textit{Hallucination Mitigation via RAG}, forcing arbitration based on real-time evidence; (2) \textit{Explainability}, as every pruning operation generates a structured log containing search evidence and reasoning; and (3) \textit{Dynamic Knowledge Adaptation}, allowing the system to correctly classify newly discovered rare diseases or drugs beyond the model's pre-training cutoff.
Finally, we established a five-tier taxonomy comprising 1,471,361 entities.
\subsection{Scalable Mapping via AC Automaton}
Post-construction, the challenge shifts to efficiently mapping the tens of millions of entries in the corpus to this established taxonomy. We implemented the Aho-Corasick (AC) Automaton algorithm~\cite{aho1975efficient}. By constructing a Trie tree with Fail Pointers from the finalized entity dictionary, we achieve $\mathcal{O}(N)$ time complexity for corpus scanning. This decouples processing speed from dictionary size, allowing us to rapidly calculate knowledge coverage and extract representative entities for massive data in data lake in approximately $20$ hours.
\subsection{Entity-Driven Data Acquisition \& Filtering}
Leveraging the constructed Medical Entity Tree (MET) as a navigation compass, as shown in bottom left part in Fig.\,\ref{fig:framework}, we established a systematic pipeline to acquire, filter, and synthesize high-quality multi-modal data. This approach shifts the paradigm from randomized collection to \textit{knowledge-guided} acquisition.
To navigate the entropy of massive raw medical data lakes, we designed a two-step retrieval and purification mechanism.
\textbf{Step 1: Node-Guided Retrieval.} Unlike traditional blind web crawling, we utilize entity nodes from the MET as high-recall query anchors to retrieve relevant images and texts from the Raw Medical Data Lake. This ensures that the retrieved corpus inherently aligns with the medical ontology distribution.
\textbf{Step 2: Two-Stage Hybrid Filtering and Alignment.} To balance filtering quality and computational cost on typical billion-scale datasets, we implemented a ``Small-Model Collaboration'' strategy: (1) \textit{MLLM Annotation (Teacher):} We randomly sample a small subset of the retrieved data and employ a powerful MLLM to perform zero-shot quality assessment and labeling. (2) \textit{ViT Distillation (Student):} These high-quality labels supervise the training of a lightweight Vision Transformer (ViT) classifier. (3) \textit{Scalable Filtering:} The trained ViT is deployed on the full dataset for rapid inference, filtering out non-medical or low-quality images. (4) \textit{Entity-Visual Alignment:} A final verification step checks the semantic consistency between the visual content and the MET entity tag (e.g., verifying that an image tagged ``Liver Cirrhosis'' indeed exhibits hepatic lesions), effectively eliminating noise.
\subsection{Knowledge-Aware Data Synthesis \& Enhancement}
This phase aims to sublimate raw data into high-quality knowledge-aware samples that foster advanced knowledge reasoning capabilities. By exploiting the structural constraints and logical pathways inherent in the Medical Entity Tree (MET), we generate task-oriented data through a dual-track synthesis pipeline.
\textbf{Track 1: Contextual Re-Captioning.} We address the common issue of raw web-scraped medical data containing noisy or sparse Alt Text. We fuse the original text with its corresponding hierarchically linked entities retrieved from the MET. A MLLM, prompted as an expert in medical imaging analysis, then synthesizes these inputs into a single, enriched, and contextualized caption. To ensure high-fidelity alignment, this generation process is strictly governed by several constraints: the model must first analyze the visual evidence to produce an objective, fact-based caption grounded entirely in the image. Furthermore, it must inject precise medical terminology derived directly from the linked entities and utilize the hierarchical context to create a highly structured description. For example, if the entities involve ``Lobar Pneumonia'' and ``Consolidation,'' the generator must explicitly explain that the consolidation is a visual feature of the pneumonia.
\textbf{Track 2: Structure-Constrained Reasoning Synthesis.} To move the model beyond simple pattern recognition toward complex causal reasoning, we generate reasoning-intensive training samples based on dynamic logical inference chains extracted from the MET, such as the progression from ``Liver Cirrhosis'' to ``Liver'' to ``CT''. The generator MLLM is conditioned to identify and analyze fine-grained visual biomarkers in the image—such as margins, texture, or intensity—that serve as the ground truth for the clinical transitions. This track produces two specific task formats: Multiple-Choice Questions (MCQs) featuring medically plausible distractors that must be definitively ruled out by visual evidence, and Interpretation tasks that require the model to judge the validity of a clinical claim derived from the inference chain. To guarantee the rigor of these reasoning samples, the synthesis enforces strict image dependency, ensuring the answer is impossible to guess from the text alone without analyzing the specific visual features. Additionally, it demands causal depth by requiring a step-by-step causal explanation that explicitly states which visual finding confirms the target diagnostic path and rules out alternative diagnoses.
For details, refer to \textbf{Supp. B}.
\section{Experiments}
\label{sec:experiments}
\subsection{Implementation Details}
Our model builds upon Qwen2.5-VL-Instruct model, with parameter
sizes of 7B, and is further optimized 2 epoch via continual training with base learning rate of $1e^{-5}$. We adopt the AdamW optimizer with the cosine learning rate scheduler and a warm-up step of 100. The maximum sequence length is set to 8,192 tokens,
the per-device training batch size is configured as 2, and the gradient accumulation step is set to 2.
Vision encoder remains frozen, only the LLM and the projection
layer are fine-tuned.
The synthetic caption data and VQA data used for training were 1.8 million and 5 million, respectively.
More implementation details refer to \textbf{Supp. C}.
\begin{table}[htbp]
\vspace{-10pt}
\centering
\caption{Semantic Coverage Analysis: Forward vs. Backward AMCS}
\label{tab:coverage_results}
\begin{adjustbox}{max width=0.98\linewidth}
\begin{tabular}{lcc}
\toprule
\textbf{Target Set ($A$)} & \textbf{Forward Coverage} & \textbf{Backward Coverage} \\
& $\text{AMCS}(A, Ref)$ & $\text{AMCS}(Ref, A)$ \\
\midrule
Clinical Knowledge Data & 0.96 & 0.68 \\
Common Crawl Medical Corpus & 0.95 & 0.89 \\
CMeKG~\cite{byambasuren2019cmekg} & 0.97 & 0.79 \\
\bottomrule
\end{tabular}
\end{adjustbox}
\vspace{-15pt}
\end{table}
\subsection{Quantitative Evaluation of Knowledge Coverage}
The core challenge in constructing a medical taxonomy lies in defining and measuring ``comprehensiveness.'' Our methodology is grounded in the hypothesis that \textit{``open-source medical literature encapsulates the totality of currently known human medical knowledge''}. Consequently, a taxonomy distilled directly from these sources serves as the ground truth. To quantitatively validate this, we propose a semantic coverage metric to compare our taxonomy against established benchmarks.
We employ \textbf{Average Max Cosine Similarity (AMCS)}, a semantic similarity metric, to measure the extent to which a Target Set ($A$) is covered by a Reference Set ($Ref$). The metric is defined as: $\textbf{AMCS}(A, Ref) = \frac{1}{|A|} \sum_{i=1}^{|A|} \max_{j \in [1, |Ref|]} \text{sim}(\mathbf{a}_i, \mathbf{r}_j)$,
where $\text{sim}(\cdot)$ is $\text{Cosine Similarity}$, $\mathbf{a}_i$ and $\mathbf{r}_j$ denote the embedding vectors of entities in set $A$ and set $Ref$, respectively, generated by a pre-trained encoder (e.g., Sentence-BERT~\cite{reimers2019sentence}). This metric is asymmetric, offering two distinct interpretations: (1) Forward Coverage $\text{AMCS}(A, Ref)$: A high value indicates that most concepts in the target set $A$ are semantically present in our Reference taxonomy. This establishes the validity of our tree. (2) Backward Coverage $\text{AMCS}(Ref, A)$: A lower value implies that our Reference taxonomy contains numerous concepts absent from set $A$.
We designate our constructed Medical Entity Tree as the Reference Set ($Ref$). We selected three distinct datasets as Target Sets ($A$) for evaluation: (1) Clinical Knowledge Data: clinical capability benchmarks that measure the coverage of various aspects
of the model, representing clinical practice focus. (2) Common Crawl Medical Corpus: A large-scale general medical dataset representing broad data distribution. (3) CMeKG (Chinese Medical Knowledge Graph)~\cite{byambasuren2019cmekg}: An authoritative open-source benchmark utilizing natural language processing on massive medical texts, referencing international standards like ICD~\cite{world2009international}.
The quantitative results are presented in Table~\ref{tab:coverage_results}.
The analysis yields two critical insights: (1) Near-Complete Forward Coverage: Our Medical Entity Tree achieves Forward Coverage AMCS scores exceeding $0.95$ across all datasets (0.96, 0.95, 0.97). This confirms that our tree successfully encompasses the vast majority of medical concepts found in both clinical benchmarks and established knowledge graphs like CMeKG~\cite{byambasuren2019cmekg}. (2) Significant Backward Gap: Conversely, the lower Backward Coverage AMCS scores (e.g., 0.68 for Clinical Knowledge Data, 0.79 for CMeKG) demonstrate that our MET contains a substantial number of long-tail entities and fine-grained concepts not present in the baselines. This empirically supports our hypothesis that extracting directly from authoritative literature yields a more comprehensive ontology than existing collections.
\subsection{Comparison on Medical Benchmarks}
\begin{table*}[htbp]
    \centering
    \vspace{-20pt}
    \caption{\textbf{Performance comparison with state-of-the-art models on six authoritative medical benchmarks.} The evaluation includes both general-purpose MLLMs and specialized medical models. Ours-7B achieves the highest average accuracy, demonstrating the effectiveness of the proposed entity-centric medical data engineering framework. (*) indicates the amount of data.}
    \label{tab:med_vqa_results}
    \begin{adjustbox}{max width=1.0\linewidth}
    \begin{tabular}{lccccccc}
        \toprule
        \textbf{Models} & \textbf{\begin{tabular}[c]{@{}c@{}}MMMU-Med~\cite{yue2024mmmu}\\ (150)\end{tabular}} & \textbf{\begin{tabular}[c]{@{}c@{}}VQA-RAD~\cite{lau2018dataset}\\ (451)\end{tabular}} & \textbf{\begin{tabular}[c]{@{}c@{}}SLAKE~\cite{liu2021slake}\\ (2094)\end{tabular}} & \textbf{\begin{tabular}[c]{@{}c@{}}PathVQA~\cite{he2020pathvqa}\\ (6719)\end{tabular}} & \textbf{\begin{tabular}[c]{@{}c@{}}PMC-VQA~\cite{zhang2023pmc}\\ (2000)\end{tabular}} & \textbf{\begin{tabular}[c]{@{}c@{}}OmniMedVQA~\cite{hu2024omnimedvqa}\\ (88996)\end{tabular}} & \textbf{Avg.} \\ \midrule
        \multicolumn{8}{c}{\textbf{General Models}} \\ \midrule
        LLaVa1.6-7B~\cite{liu2023visual} & 36.00 & 49.22 & 45.51 & 29.57 & 36.40 & 50.67 & 41.23 \\
        Qwen2.5-VL-7B~\cite{wang2024qwen2} & 56.00 & 62.08 & 61.41 & 38.80 & 53.65 & 63.42 & 55.89 \\
        InternVL3-8B~\cite{chen2024internvl} & 58.00 & 61.64 & 70.73 & 40.79 & 53.95 & 78.45 & 60.59 \\ \midrule
        \multicolumn{8}{c}{\textbf{Medical Models}} \\ \midrule
        LLaVa-Med-7B~\cite{li2023llava} & 29.30 & 53.70 & 48.00 & 38.80 & 30.50 & 44.30 & 40.77 \\
        BioMediX2-8B~\cite{mullappilly2024bimedix2} & 39.80 & 49.20 & 57.70 & 37.00 & 43.50 & 63.30 & 48.42 \\
        HuatuoGPT-V-7B~\cite{chen2024towards} & 47.30 & 67.00 & 67.80 & 48.00 & 53.30 & 74.20 & 59.60 \\
        Lingshu-7B~\cite{xu2025lingshu} & 54.00 & 67.90 & \textbf{83.10} & \textbf{61.90} & 56.30 & 82.90 & 67.68 \\ 
        Ours-7B & \textbf{73.77} & \textbf{77.16} & 70.39 & 46.50 & \textbf{63.75} & \textbf{83.36} & \textbf{69.16} \\ 
        \bottomrule
    \end{tabular}%
    \end{adjustbox}
    \vspace{-15pt}
\end{table*}
\begin{table*}[t]
    \centering
    \caption{\textbf{Data Generalization Analysis.} We evaluate the effectiveness of our synthesized data by fine-tuning different general-purpose MLLM backbones. The results demonstrate that our entity-centric data significantly enhances medical capabilities across diverse architectures.}
    \label{tab:data_generalization}
    \begin{adjustbox}{max width=1.0\linewidth}
        \begin{tabular}{l|cccccc|c}
            \toprule
            \textbf{Base Model} & \textbf{MMMU-Med} & \textbf{VQA-RAD} & \textbf{SLAKE} & \textbf{PathVQA} & \textbf{PMC-VQA} & \textbf{OmniMedVQA} & \textbf{Avg.} \\
            \midrule
            \textit{LLaVa1.6-7B} & 36.00 & 49.22 & 45.51 & 29.57 & 36.40 & 50.67 & 41.23 \\
            \rowcolor{gray!10} \textbf{+ Our Data} & \textbf{36.67} & \textbf{56.10} & \textbf{55.44} & \textbf{34.74} & \textbf{45.05} & \textbf{58.16} & \textbf{47.69} \\
            \midrule
            \textit{InternVL3-8B} & 58.00 & 61.64 & 70.73 & 40.79 & 53.95 & 78.45 & 60.59 \\
            \rowcolor{gray!10} \textbf{+ Our Data} & \textbf{65.33} & \textbf{62.97} & \textbf{71.21} & \textbf{43.37} & \textbf{56.55} & \textbf{81.56} & \textbf{63.50} \\
            \bottomrule
        \end{tabular}
    \end{adjustbox}
    \vspace{-15pt}
\end{table*}
To rigorously evaluate the effectiveness of our proposed Entity-Centric Medical Data Engineering framework, we compared our model (Ours-7B) against a diverse set of state-of-the-art baselines, including general-purpose MLLMs such as Qwen2.5-VL-7B~\cite{wang2024qwen2}, LLaVa1.6-7B~\cite{liu2023visual} and InternVL3-8B~\cite{chen2024internvl}, as well as domain-specific medical MLLMs like Lingshu-7B~\cite{xu2025lingshu} and HuatuoGPT-V-7B~\cite{chen2024towards}. As summarized in Table~\ref{tab:med_vqa_results}, our model delivers robust performance across the evaluated spectrum, achieving a leading average score of 69.16\%, which surpasses the previous state-of-the-art medical model, Lingshu-7B~\cite{xu2025lingshu} (67.68\%), and significantly outperforms robust generalist models like InternVL3-8B~\cite{chen2024internvl} (60.59\%). This quantitative advantage underscores the efficacy of anchoring multimodal data to a structured Medical Entity Tree (MET) rather than relying on conventional, coarse-grained data partitioning.
Our model exhibits particularly strong capabilities in tasks requiring expert-level reasoning and fine-grained knowledge application. Notably, on the MMMU-Med~\cite{yue2024mmmu}, which is designed to test multidisciplinary reasoning and expert AGI capabilities, our approach achieves a score of 73.77\%, establishing a substantial margin over both Qwen2.5-VL-7B~\cite{wang2024qwen2} (56.00\%) and Lingshu-7B~\cite{xu2025lingshu} (54.00\%). Furthermore, we observe consistent improvements on large-scale datasets such as OmniMedVQA~\cite{hu2024omnimedvqa} and PMC-VQA~\cite{zhang2023pmc}, where our model scores 83.36\% and 63.75\% respectively, outperforming the strongest baselines in these benchmarks. These results suggest that our synthesized data effectively mitigates the fragmentation issues inherent in traditional data silos and empowers the model to internalize complex medical interconnections.
While prior models such as HuatuoGPT-V~\cite{chen2024towards} and Lingshu~\cite{xu2025lingshu} have demonstrated competitive performance on specific benchmarks like SLAKE~\cite{liu2021slake} and PathVQA~\cite{he2020pathvqa}, our model maintains a superior balance across the full range of diagnostic modalities and clinical tasks. The significant gains in the ``Average'' metric indicate that the Entity-Centric Medical Data Engineering framework successfully facilitates a systematic pedagogical path similar to human medical expertise, enabling the model to generalize better across diverse medical contexts ranging from radiology to pathology. By treating entities as the fundamental unit of data organization, our method ensures high-fidelity alignment between visual features and medical concepts, resulting in a more comprehensive and robust medical MLLM.
\subsection{Data Generalization}
\begin{figure*}[!t]
    \centering
    \begin{subfigure}[b]{0.49\linewidth}
        \centering
        \includegraphics[width=\linewidth]{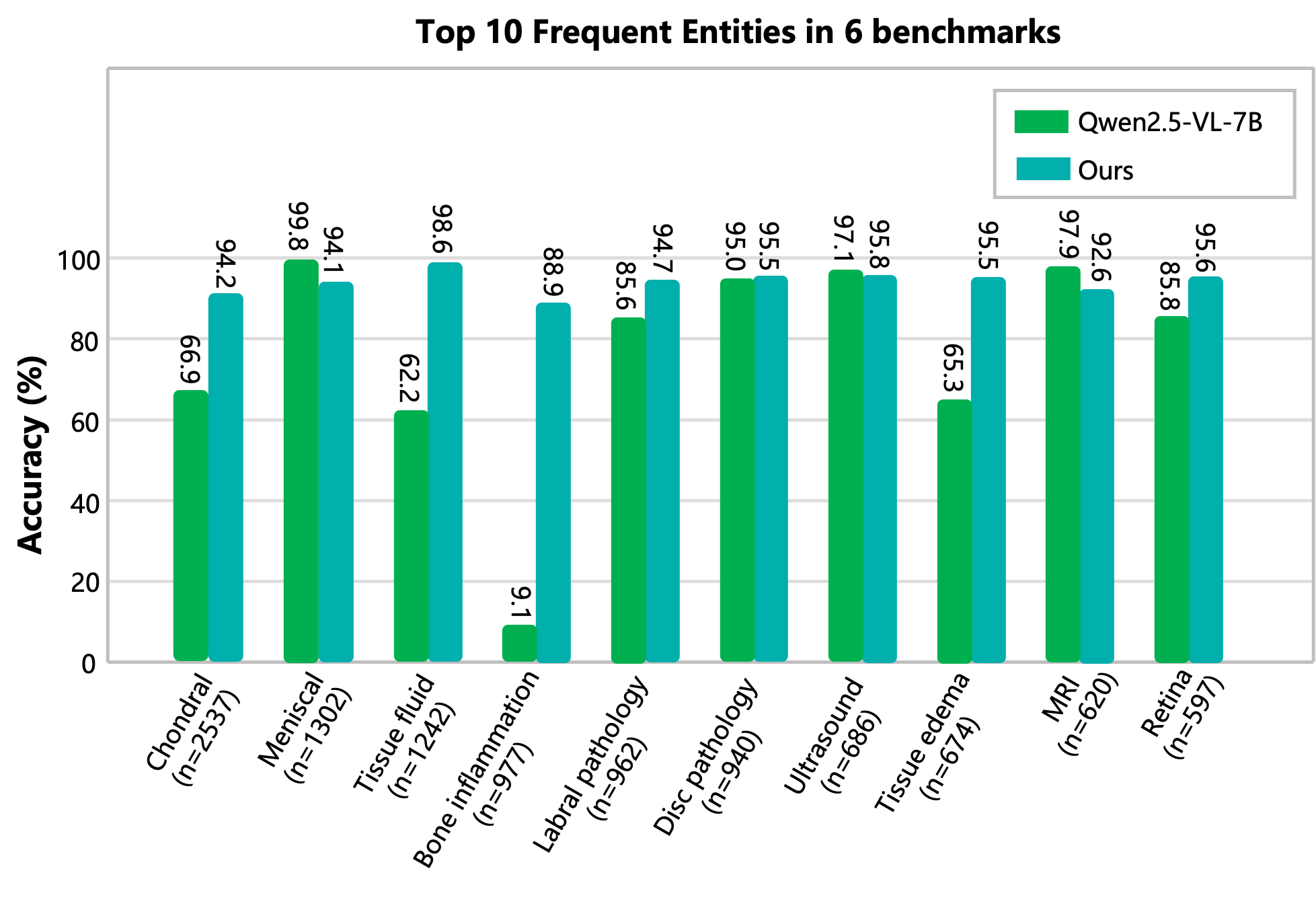}
        \caption{Top 10 Frequent Entities in 6 benchmarks}
        \label{fig:accuracy_top10}
    \end{subfigure}
    \hfill
    \begin{subfigure}[b]{0.49\linewidth}
        \centering
        \includegraphics[width=\linewidth]{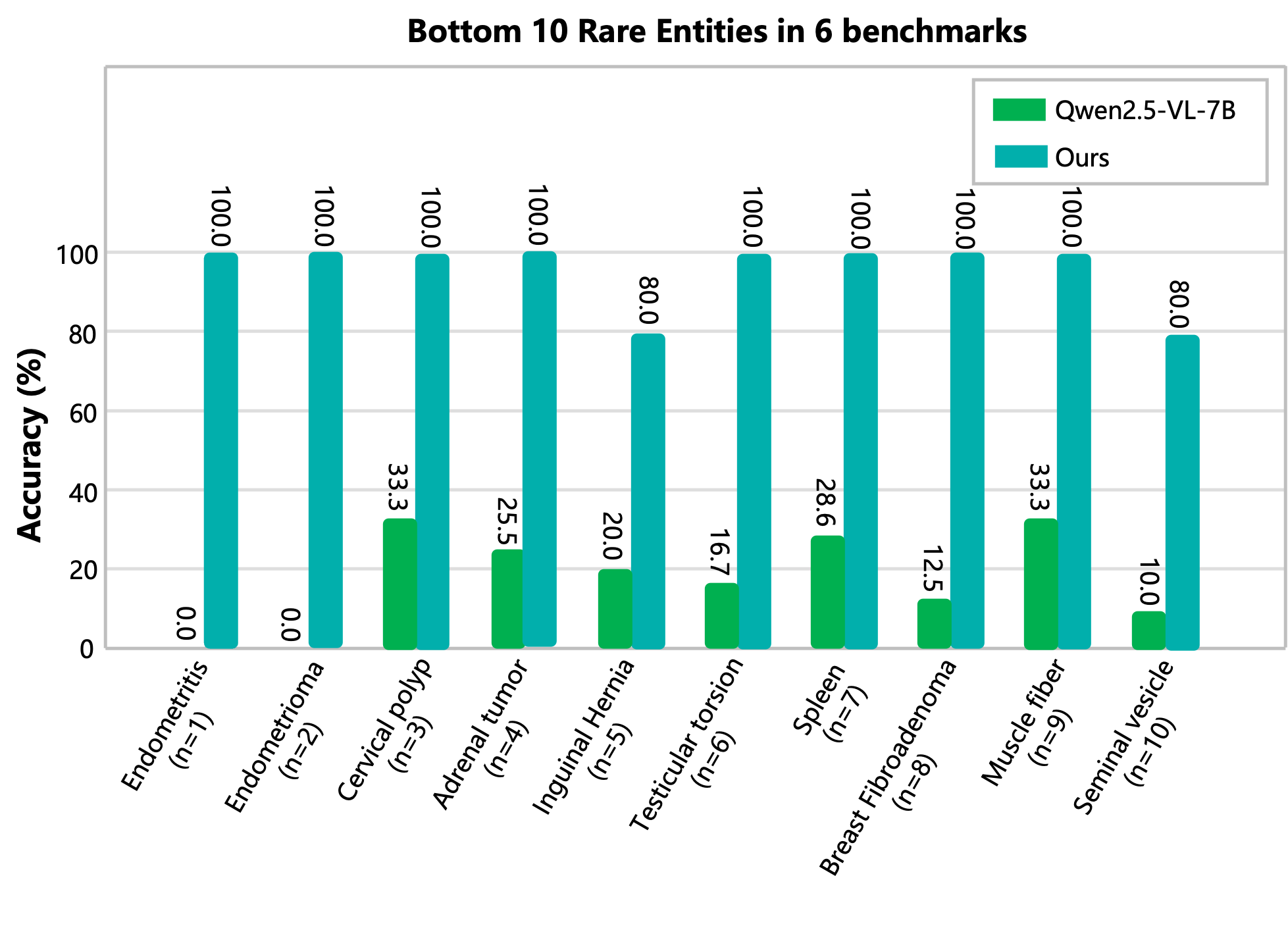}
        \caption{Bottom 10 Rare Entities in 6 benchmarks}
        \label{fig:accuracy_bottom10}
    \end{subfigure}

    \caption{\textbf{Granular performance analysis on frequent and rare medical entities within the evaluated benchmarks.} (a) Accuracy comparison on the top 10 most frequent entities, where our framework significantly boosts performance in challenging entities like bone inflammation and tissue fluid while maintaining competitive levels in saturated ones. (b) Accuracy comparison on the bottom 10 rare entities, demonstrating our model's superior coverage of the data distribution's long tail. Our method achieves near-perfect recognition for specific conditions where the general-purpose baseline often fails completely (0\% accuracy) due to data sparsity and safety-related knowledge gaps.}
    \label{fig:accuracy_comparison_all_datasets}
    \vspace{-15pt}
\end{figure*}
To assess the scalability and unique value of our Entity-Centric Medical Data Engineering framework, we conducted generalization experiments by fine-tuning two representative general-purpose models, LLaVa1.6-7B~\cite{liu2023visual} and InternVL3-8B~\cite{chen2024internvl}, with our synthesized, entity-centric data.
As shown in Table \ref{tab:data_generalization}, integrating our data yields consistent performance gains across all benchmarks. For instance, LLaVa1.6-7B~\cite{liu2023visual} achieves a significant improvement of 6.46\% in average accuracy, demonstrating that our structured knowledge injection effectively mitigates the domain gap for generalist models. Similarly, the performance of the stronger InternVL3-8B~\cite{chen2024internvl} is further boosted to 63.50\%, validating that our knowledge-aligned data serves as a universal enhancer for medical multimodal learning, independent of the underlying model architecture.
The greater performance improvement in Qwen2.5-VL-7B~\cite{wang2024qwen2} may be due to its stronger generalization capabilities, allowing it to better adapt to and leverage the structured medical knowledge in our data. This cross-domain adaptability results in a more significant performance boost.
\subsection{Quantitative assessment of medical knowledge}
To provide a granular understanding of how our Entity-Centric Medical Data Engineering framework enhances clinical competency, we conducted a fine-grained performance analysis based on specific medical entities. 
This analysis evaluates the model’s discriminative capability when encountering entities that are sparsely represented across the six aggregated benchmarks, moving beyond overall scores. We compared the accuracy of our model against the Qwen2.5-VL-7B~\cite{wang2024qwen2} baseline across both the most frequent and the most rare entities found in the test sets.
The results for benchmark-rare entities (Fig\,\ref{fig:accuracy_bottom10}) reveal a significant ``knowledge void'' in general-purpose models. Notably, the baseline model exhibits a near-total inability to recognize entities such as \textit{Endometritis}, \textit{Endometrioma}, and \textit{Seminal vesicle}, frequently yielding 0\% accuracy. 
This failure is primarily attributed to the intersection of data sparsity within the benchmarks and the stringent safety alignment protocols inherent in generalist MLLMs.
Such filters, designed to restrict sensitive or explicit content, often inadvertently over-filter or suppress professional multimodal data related to reproductive organs and private anatomy. 
By anchoring our data engine to the Medical Entity Tree (MET), we effectively bypass these heuristic-based limitations, ensuring that crucial but sensitive diagnostic concepts are preserved and accurately aligned with professional visual evidence.
This allows our model to achieve 100\% accuracy on entities like \textit{Cervical polyp} and \textit{Testicular torsion}, where professional medical interpretation must prevail over generalized safety-induced suppression.
In the analysis of frequent entities (Fig\,\ref{fig:accuracy_top10}), our model demonstrates substantial performance leaps in entities requiring expert-level visual discrimination, such as \textit{Bone inflammation} (from 9.1\% to 88.9\%) and \textit{Tissue fluid} (from 62.2\% to 98.6\%).
However, for certain entities like \textit{Meniscal}, \textit{MRI}, and \textit{Ultrasound}, our method shows no significant improvement or even a marginal fluctuation compared to the baseline. This stagnation occurs because the baseline already achieves near-saturated performance (above 95\%) on these broad modalities and common findings.
Furthermore, while generalist models are often heavily biased toward recognizing frequent normal appearances or coarse modality features, our entity-driven alignment shifts the focus toward distinguishing specific pathological nuances. This suggests that for entities where general-purpose perception is already robust, the primary value of our framework is not in reinforcing these saturated concepts, but in bridging the critical gap for complex, sparsely represented diagnostic findings.
\begin{figure*}[!t]
    \centering
    \begin{subfigure}[b]{0.49\linewidth}
        \centering
        \includegraphics[width=\linewidth]{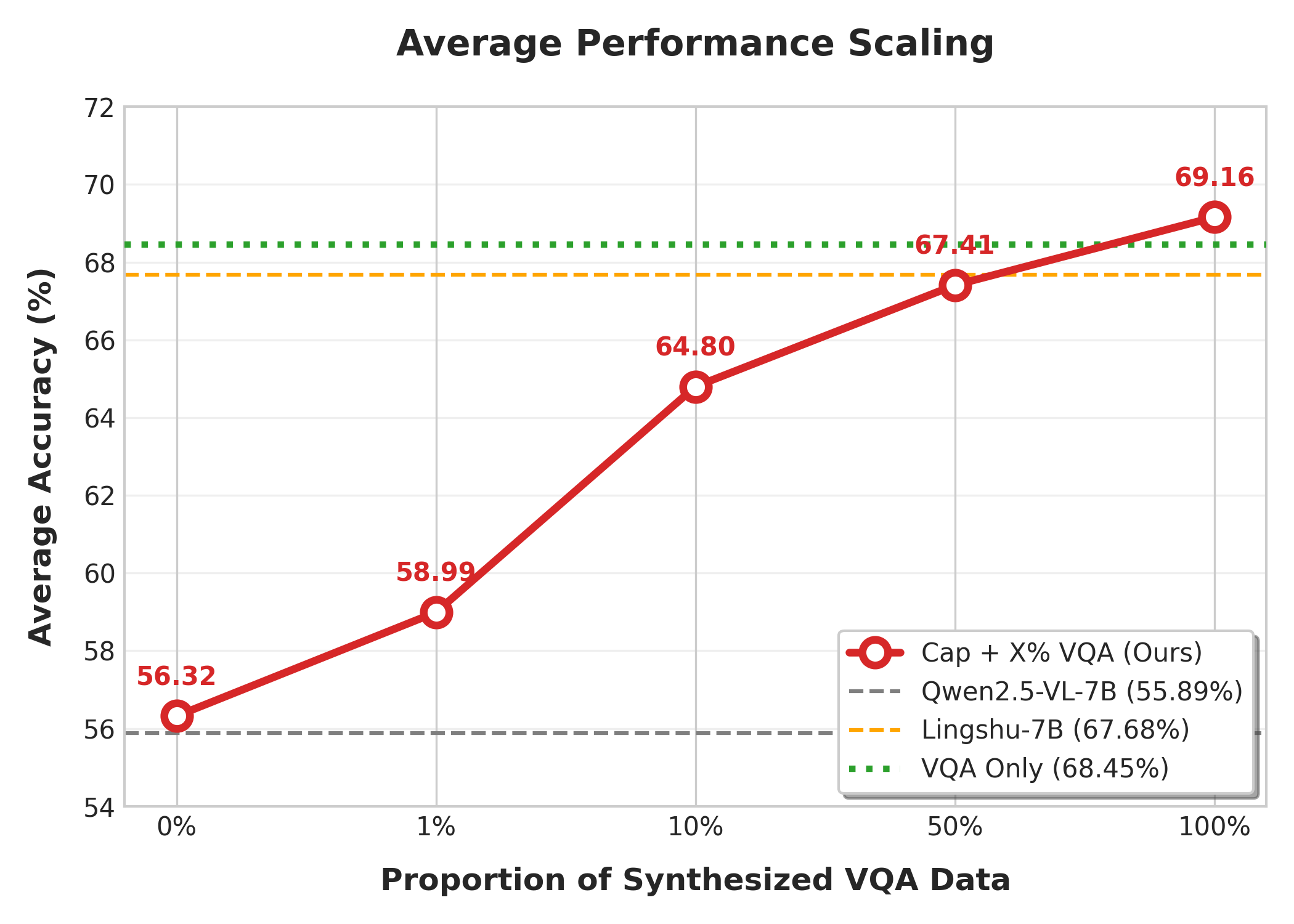}
        \caption{Average Performance Scaling.}
        \label{fig:data_scaling}
    \end{subfigure}
    \hfill
    \begin{subfigure}[b]{0.49\linewidth}
        \centering
        \includegraphics[width=\linewidth]{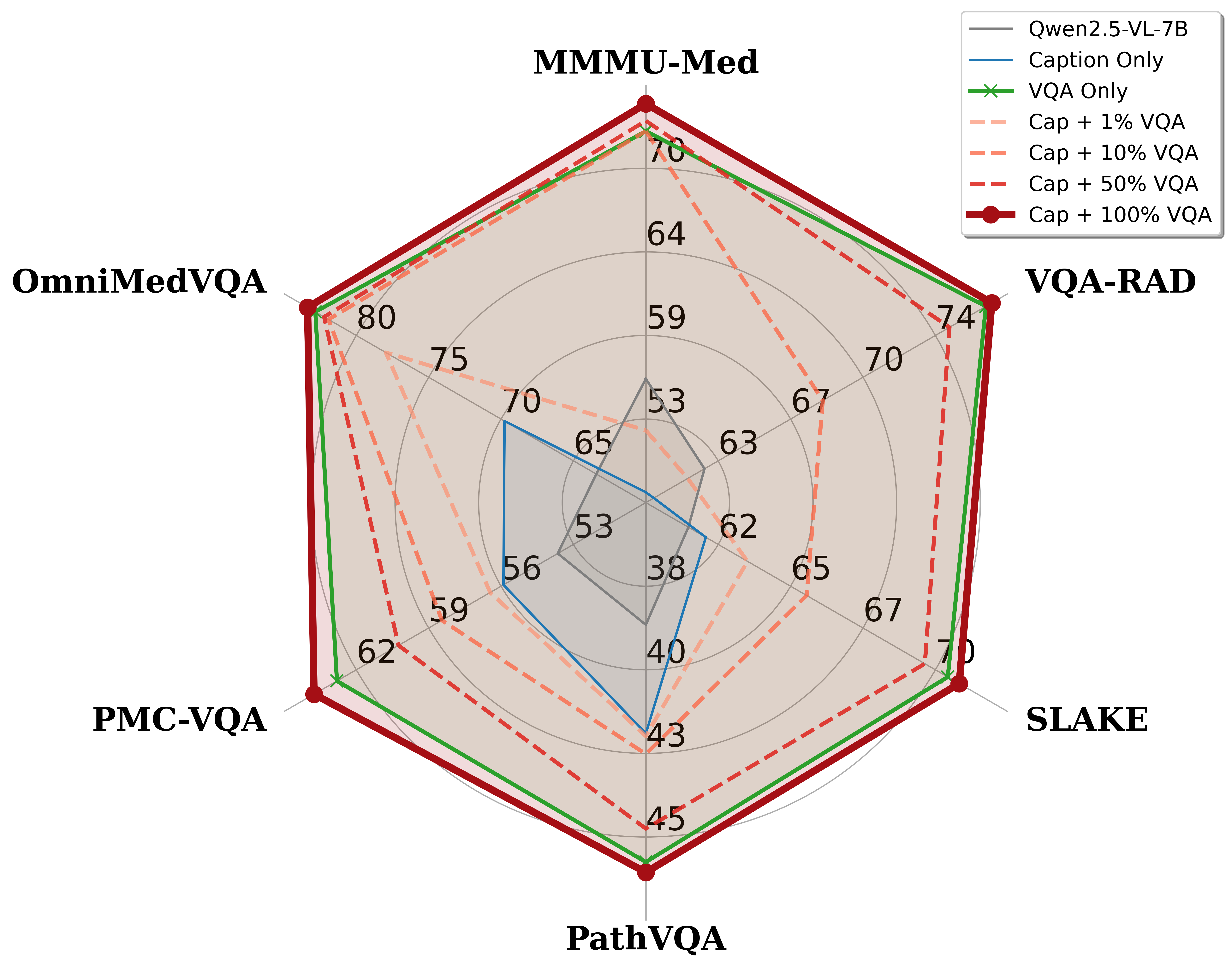}
        \caption{Benchmark Performance Comparison.}
        \label{fig:data_radar}
    \end{subfigure}

    \caption{Ablation study on the synergistic impact of the dual-track data synthesis pipeline. (a) Performance scaling as reasoning-intensive VQA data is integrated into the caption-only baseline; a 10\% inclusion of VQA data already yields a significant leap. (b) Radar chart visualizing the performance across six representative medical benchmarks. The results reveal that while VQA-driven supervision (green) establishes the necessary diagnostic logic, its integration with entity-enriched captions (dark red) provides a vital semantic foundation. }
    \label{fig:scaling_radar}
    \vspace{-15pt}
\end{figure*}
\subsection{Synergy of Declarative Knowledge and Interrogative Reasoning}
To decouple the distinct contributions of declarative knowledge and interrogative reasoning within our framework, we conduct a granular ablation study across varying data compositions. We investigate the isolated and synergistic impacts of our dual-track pipeline by comparing models trained on \textit{Track 1} (providing dense anatomical and pathological descriptions) against \textit{Track 2} (providing causal logical paths).
As illustrated in Fig\,\ref{fig:data_scaling}, training the baseline model exclusively on the re-captioned dataset yields a performance of $56.32\%$. While this establishes a fundamental visual-semantic alignment, declarative captions alone are insufficient for tasks requiring complex clinical query responsiveness. Conversely, a ``VQA Only'' configuration ($68.45\%$) significantly enhances diagnostic proficiency by teaching the model instructional logic. However, as shown in Fig\,\ref{fig:data_radar}, this approach lacks the dense semantic context provided by the Medical Entity Tree (MET), preventing it from reaching peak performance on multidisciplinary tasks.
A key insight from our scaling analysis is the non-linear catalytic effect of VQA data. We observe that introducing a nominal $10\%$ of VQA data into the caption-based training set triggers a leap to $64.80\%$. This synergistic effect suggests that our entity-enriched captions act as a \textit{latent knowledge reservoir}: the model internalizes the vast medical concepts from the MET during caption training, but requires a critical mass of structure-constrained reasoning samples to unlock the ability to format this knowledge into valid clinical responses.
This catalytic phenomenon is most pronounced when comparing benchmarks of varying complexity. On the reasoning-heavy MMMU-Med, which is designed to test multidisciplinary expert AGI capabilities , performance surges from $48.67\%$ (caption-only) to $72.00\%$ with just $10\%$ VQA data. This disproportionate jump occurs because the declarative knowledge from the MET acts as a catalyst, providing the high-level medical context necessary to solve complex, cross-disciplinary problems that logic-only training cannot reach. In contrast, on saturated benchmarks like VQA-RAD, the improvement is more linear. Because VQA-RAD focuses on standard radiological signs where general-purpose perception is already relatively robust , the primary value of our framework shifts from bridging knowledge voids to reinforcing specific pathological nuances.
Ultimately, the optimal configuration utilizing both the full caption set and the full VQA set achieves the state-of-the-art average score of $69.16\%$. These findings confirm that combining the broad semantic coverage of entity-enriched captions with the causal logical paths of VQA creates an essential complementary effect, enabling the model to generalize across diverse medical contexts ranging from routine radiology to expert-level pathology.
\section{Conclusion}
\label{sec:conclusion}
In this paper, we present a novel Entity-Centric Medical Data Engineering framework that overcomes the limitations of traditional, coarse-grained medical data curation.
By extracting entities from authoritative literature, we construct a hierarchical Medical Entity Tree (MET) as a systematic knowledge repository.
Our framework leverages the MET to align raw multimodal data through node-guided retrieval and hybrid filtering, followed by a knowledge-aware synthesis strategy that generates enriched captions and reasoning-intensive VQA pairs.
Evaluations across six medical benchmarks show that our approach significantly enhances clinical reasoning and fine-grained recognition in MLLMs, achieving SOTA performance. This work provides a scalable paradigm for empowering medical models with structured, textbook-level knowledge.
Furthermore, this approach can be extended to other verticals such as law, education, and finance, where structured knowledge is crucial. By systematically organizing domain-specific knowledge, this framework enables more accurate and context-aware reasoning across diverse fields that rely on well-defined, quantifiable knowledge.



%
%
\bibliographystyle{splncs04}
\bibliography{main}
\newpage
\appendix

 


\renewcommand{\thefigure}{\Roman{figure}}







\section{Medical Entity Tree Construction Details}
\label{supp:MET_prompt}
\begin{table*}[htbp]
\vspace{-20pt}
\begin{adjustbox}{max width=1.0\linewidth}
\begin{tcolorbox}[title=\textbf{Prompt for Stage 1: Batch Entity Extraction}]
\small
Below are several sentences. Analyze them for \textbf{medical} entity nouns and output according to the following requirements:
\begin{enumerate}
    \item Determine if it is a medical entity; if not, do not output.
    \item Separate entity nouns with commas; do not include duplicates.
    \item If there are no medical-related entity nouns, output "None".
    \item Output strictly in JSON format. The example format is as follows: \\
    \textbf{\{'Sentence0': 'Entity1,Entity2,...',\\
    'Sentence1': 'Entity1,Entity2,...', ...\}}
\end{enumerate}
Sentences:\\
\{lines\}
\end{tcolorbox}
\end{adjustbox}
\vspace{-20pt}
\end{table*}
\begin{table*}[htbp]
\begin{adjustbox}{max width=1.0\linewidth}
\begin{tcolorbox}[title=\textbf{Prompt for Stage 2: Joint Extraction and Typing}]
\small
Below are several sentences. Analyze these sentences for \textbf{medical} entity nouns and their types, and output according to the following requirements:
\begin{enumerate}
    \item Entity nouns must be informative proper nouns. Secondarily, determine if they are \textbf{medical} entities; if not, do not output.
    \item Pay attention to overly long medical entity nouns and determine if they can be segmented/split.
    \item The sentences below may contain special symbols and meaningless spaces; please ignore them directly.
    \item Replace \textless EntityType\textgreater\ with the specific entity category.
    \item Replace \textless EntityName\textgreater\ with the specific entity noun.
    \item Output strictly in JSON format. The example format is as follows: \\
    \textbf{\{
        'Sentence0': [\textless EntityType\textgreater:\textless EntityName\textgreater, ...], 
        'Sentence1': [...], 
        ...
    \}}
\end{enumerate}
Sentences:\\
\{lines\}
\end{tcolorbox}
\end{adjustbox}
\vspace{-20pt}
\end{table*}
The construction of the Medical Entity Tree (MET) is a crucial step in our framework, designed to support data retrieval, alignment, and synthesis for Multimodal Large Language Models (MLLMs). The MET serves as a structured medical knowledge repository, which enables the systematic organization and processing of medical data across different modalities. The pipeline for constructing the MET is divided into three stages: entity extraction, entity typing and clustering, and tree attachment and conflict resolution.
In the \textbf{Stage 1: High-Efficiency Entity Extraction via Prompt Engineering}, as shown in {\color{green}Prompt for Stage 1}, entity extraction is performed using Large Language Models (LLMs) to identify relevant medical entities from a wide range of medical literature, including textbooks and research papers. The extracted entities—such as diseases, anatomical structures, symptoms, and modalities—are carefully filtered to exclude non-medical terms. This stage produces a large set of medical entities that form the foundation for further refinement and structuring in the stage 3.
The \textbf{Stage 2: Joint Typing and Hierarchical Clustering} involves entity typing (see {\color{green}Prompt for Stage 2}) and hierarchical clustering, where the raw entities are categorized into specific medical types (e.g., diseases, symptoms, anatomical parts). Using clustering techniques such as K-Means, entities are grouped based on their semantic relationships. This step builds a hierarchical structure that reflects the interconnected nature of medical knowledge, organizing entities into parent-child relationships. The structured set of entities generated in this stage is crucial for the subsequent alignment and retrieval tasks. This stage provides a core taxonomy for stage 3.
The \textbf{Stage 3: Controlled Expansion via Tree Attachment} focuses on tree attachment (see {\color{green}Prompt for Stage 3}) and conflict resolution (see {\color{green}Prompt for Stage 3 Agent}), where the structured entities are integrated into the existing Medical Entity Tree. Each entity is inserted into its appropriate category within the tree, ensuring that the hierarchical relationships established in the previous stages are maintained. When an entity could fit into multiple parent categories, a conflict resolution agent is employed to ensure the entity is placed in the most appropriate node based on medical principles such as etiological dominance and specificity. This ensures that the MET remains coherent and scalable.
Once the MET is constructed, it acts as the foundation for knowledge-driven data tasks. The MET is used for data retrieval, where relevant data is queried based on the entities within the tree. It also facilitates data alignment, ensuring that the extracted data from different sources matches the corresponding entities and relationships. Finally, the MET supports data synthesis, where the hierarchical structure of the tree is leveraged to generate enriched medical data, such as structured captions and reasoning-intensive Q\&A pairs. This systematic, entity-driven approach ensures high-quality, structured medical data for training MLLMs and supporting advanced medical reasoning tasks.
\begin{table*}[htbp]
\begin{adjustbox}{max width=1.0\linewidth}
\begin{tcolorbox}[title=\textbf{Prompt for Stage 3: Entity Tree Attachment}]
\small
You are a medical entity taxonomy expert. You must strictly follow the guidelines below to integrate the medical entity noun I provide into the existing medical entity tree.

\textbf{Maintain Original Structure}:\\
You are NOT allowed to change the names of nodes on the medical entity tree.

\textbf{Precise Insertion}:\\
Insert the medical entity noun into the appropriate sub-categories of its parent node. Ensure only valid medical entity is inserted and adhere to hierarchical relationships.

\textbf{Rationality of New Insertions}:\\
If you insert a medical entity into the tree, you must provide the reason for its insertion to ensure interpretability and traceability.\\
Output Format: \textless Reason\textgreater xxx \textless /Reason\textgreater \\
\textless InsertionPath\textgreater Node1.Node2.InsertedNode\textless /InsertionPath\textgreater

\textbf{Handling Unclassifiable Cases}:\\
Report medical entity noun that cannot be classified and explain the reasons for the uncertainty. If unsure about the classification, provide detailed reasoning.\\
Output Format: \textless Reason\textgreater xxx\textless /Reason\textgreater \textless Reasoning\textgreater yyy\textless /Reasoning\textgreater

Medical Entity Tree:\\
\{tree\}

Medical Entity Noun:\\
\{entity\}
\end{tcolorbox}
\end{adjustbox}
\end{table*}
\section{Knowledge-Aware Data Synthesis \& Enhancement Details}
\label{supp:data_prompt}
The knowledge-aware data synthesis and enhancement pipeline is designed to convert raw medical data into high-quality, task-oriented training samples that support both visual understanding and reasoning for Multimodal Large Language Models (MLLMs). This process leverages the hierarchical structure of the Medical Entity Tree (MET) to ensure that synthesized data is semantically precise, visually grounded, and causally coherent.
The pipeline operates in two complementary tracks. In Track 1 (see {\color{green}Prompt for Track 1}), Contextual Re-Captioning, each medical image is paired with an original caption—often noisy or sparse from web sources—and a set of hierarchically linked medical entities retrieved from the MET. This track fuses the visual evidence in the image with the semantic knowledge of the linked entities and the information in the original caption to produce a enriched and structured description. The hierarchical context of the entities ensures that the caption accurately reflects the relationships between diseases, anatomical structures, and clinical signs, while preserving objectivity and grounding in the visual evidence.
In Track 2 (see {\color{green}Prompt for Track 2}), Structure-Constrained Reasoning Synthesis, the focus shifts from descriptive annotation to reasoning-intensive sample generation. Each image is associated with a dynamic inference chain derived from the MET, representing the causal and hierarchical relationships among medical entities. This track generates multiple-choice questions or interpretation tasks that require the model to perform causal reasoning based on fine-grained visual biomarkers. By linking visual findings to specific entities in the inference chain, this track enforces image-dependent, stepwise explanations that teach the model to justify clinical conclusions and rule out alternative diagnoses.
The two tracks work synergistically: Track 1 supplies declarative, entity-enriched captions that provide broad semantic context, while Track 2 adds interrogative reasoning tasks that enforce causal and stepwise analysis. This combination enables the model to internalize hierarchical medical knowledge from the MET while practicing structured reasoning.
\begin{table*}[htbp]
\begin{adjustbox}{max width=1.0\linewidth}
\begin{tcolorbox}[title=\textbf{Prompt for Stage 3 Agent: Conflict Resolution}]
\small
You are a rigorous medical knowledge base construction Agent. Your task is to resolve entity classification conflicts.
The current entity ``\{entity\}'' is attached to multiple parent nodes:

1. Parent Path A: \{path\_a\}
2. Parent Path B: \{path\_b\}
...

You must search for the exact medical definition via Google/Wiki and adjudicate based on the following principles:
\begin{enumerate}
    \item \textbf{Principle of Etiological Dominance}: Classification based on pathological mechanism or anatomical location takes precedence over clinical symptoms.
    \item \textbf{Principle of Specificity}: If one parent is a subset of another and describes the entity more accurately, prefer the more specific one.
\end{enumerate}

Thinking Steps:

Step 1: Construct query terms and call tools to search for the definition.

Step 2: Analyze results and compare the validity of each parent path.

Step 3: Decide which path to keep and delete the others.

Output Format: \\
\textless SearchEvidence\textgreater Excerpt from Wiki/Search...\textless /SearchEvidence\textgreater \\
\textless Reasoning\textgreater Since evidence shows..., and Path A focuses on XX while Path B focuses on XX, according to the Principle of Etiological Dominance, Path A is more appropriate.\textless /Reasoning\textgreater \\
\textless FinalAction\textgreater Keep: path\_a, Delete: path\_b\textless /FinalAction\textgreater
\end{tcolorbox}
\end{adjustbox}
\end{table*}
\begin{table*}[htbp]
\begin{adjustbox}{max width=1.0\linewidth}
\begin{tcolorbox}[title=\textbf{Prompt for Track 1: Contextual Re-Captioning}]
\small
You are an AI expert in medical imaging analysis and contextual captioning. Your task is to perform \textbf{“Contextual Re-Captioning”}.

You will be given:
\begin{enumerate}
    \item A medical image.
    \item An original\_caption (like a noisy or sparse Alt Text from the web).
    \item A set of hierarchically medical linked\_entities relevant to the image.
\end{enumerate}
Your goal is to \textbf{synthesize} these inputs into a single, enriched, and contextualized caption.

\textbf{Instructions:}
\begin{enumerate}
    \item First, analyze the visual evidence in the medical image.
    \item Review the original\_caption to understand its starting point, even if it is noisy or sparse.
    \item Your main task is to \textbf{fuse} the original\_caption with the linked\_entities and your visual analysis.
    \item \textbf{Inject} the precise medical terminology from the linked\_entities into a new, comprehensive description.
    \item Use the \textbf{hierarchical context} of the entities to create a more structured and informative description. For example, if an entity is “Lobar Pneumonia” and another is “Consolidation,” explain that the consolidation is a feature of the pneumonia.
    \item The final caption must be objective, fact-based, and grounded in the visual evidence.
\end{enumerate}
\textbf{Output requirements:}
\begin{enumerate}
    \item Write in English.
    \item Produce a single, detailed, and coherent paragraph.
    \item The output should be the final enriched caption \textbf{ONLY}.
\end{enumerate}

original\_caption: \{original\_caption\} \\
linked\_entities: \{entities\}
\end{tcolorbox}
\end{adjustbox}
\end{table*}
\begin{table*}[htbp]
\begin{adjustbox}{max width=1.0\linewidth}
\begin{tcolorbox}[title=\textbf{Prompt for Track 2: Structure-Constrained Reasoning Synthesis}]
\small
You are an expert medical AI specializing in knowledge-driven data synthesis. Your task is to generate reasoning-intensive training samples (Multiple-Choice Questions or Interpretation tasks) based on a provided medical image and a dynamic chain of related entities.
I will give you: A medical image \& A specific \textbf{inference\_chain} consisting of associated medical entities and their logical relationships (e.g., specific diseases, anatomical sites, or clinical signs).

\textbf{Your Overall Goal:} \\
Construct a question that moves beyond simple pattern recognition toward \textbf{CAUSAL REASONING}. The question must force the model to analyze visual biomarkers in the image to justify why the specific entities in the \textbf{inference\_chain} are present or linked, often by ruling out differential diagnoses.

\textbf{======================} \\
\textbf{STEP 1: Analyze the Inference Chain} \\
Analyze the specific entities and hierarchical transitions provided in the \textbf{inference\_chain}. Identify the core clinical knowledge point and how the entities connect (e.g., how a specific sign leads to a specific diagnosis).

\textbf{STEP 2: Inspect the Image for Visual Evidence} \\
Identify the fine-grained visual features (e.g., margins, texture, intensity) in the image that serve as the "ground truth" for the transitions in the \textbf{inference\_chain}.

\textbf{STEP 3: Generate the Question (MCQ or Judgement)} \\
Generate ONE of the following types as specified by the task:
\begin{enumerate}
    \item \textbf{MULTIPLE-CHOICE QUESTION (MCQ)}: Provide 4 options (A, B, C, D). Distractors must be medically plausible alternatives that could be confused with the target path but are ruled out by visual evidence.
    \item \textbf{INTERPRETATION / JUDGEMENT}: Formulate a clinical claim derived from the \textbf{inference\_chain}. The task is to judge its validity (True/False or Yes/No) based on the image.
\end{enumerate}

\textbf{OUTPUT FORMAT} \\
Return your final answer in strict JSON format:
\tiny
\begin{verbatim}
{
  "target_knowledge_path": "{inference_chain}",
  "question_type": "<'MCQ' or 'Interpretation'>",
  "vqa_data": {
    "question": "<The reasoning-based question or claim>",
    "options": ["A)...", "B)...", "C)...", "D)..."] (null if Interpretation),
    "correct_answer": "<Letter or True/False>",
    "explanation": "A step-by-step causal explanation linking visual signs to the entities."
  }
}
\end{verbatim}
\small
\textbf{Constraints:} \\
\textbf{Verify Image Dependency}: The answer must be impossible to guess from the text alone without analyzing the specific visual features. \\
\textbf{Causal Depth}: The explanation must explicitly state which visual finding confirms the target path and rules out alternatives. \\
Never output comments or markdown, only output a valid JSON object.

\textbf{Entities \& Inference Chain:} \\
\{inference\_chain\}
\end{tcolorbox}
\end{adjustbox}
\end{table*}
\section{More Implementation Details}
\label{supp:more_implementation_details}
\subsection{Prevent Benchmarks Leakage}
To address the issue of benchmarks leakage and near-duplicate contamination, we utilized a two-pronged approach combining hash filtering and CLIP-based image similarity filtering.

\textbf{Hash Filtering:} We used perceptual hashing techniques to ensure that identical or near-duplicate images were excluded from the training and test sets. This method helps to identify and remove duplicates by generating unique hash codes for each image. If any two images had identical or similar hash values, they were deemed as duplicates, effectively reducing redundancy and preventing leakage.

\textbf{CLIP Image Similarity Filtering:} In addition to hash filtering, we also employed CLIP~\cite{radford2021learning} to evaluate image similarity. By extracting features using CLIP’s powerful visual encoder, we compared the similarity of image pairs. If two images were too similar, they were filtered out to avoid contamination due to near-duplicate content.
\subsection{Entity–Visual Alignment Verification}
We implemented a two-stage approach to ensure high-quality alignment between entities and their corresponding visual features:

\textbf{First Stage: CLIP-based Coarse Filtering.} In the first stage, we use the CLIP~\cite{radford2021learning} for coarse filtering. CLIP is employed to compute the similarity between the textual entity descriptions and the visual features of the images. For each image and its corresponding entity, we calculate the cosine similarity between the image's visual embeddings and the entity's textual embeddings. The images with a similarity score below a pre-defined threshold are discarded, as they are considered to have weak or no alignment with the given entity. We set the threshold $0.3$ for cosine similarity based on empirical evaluation. This stage may fail in cases of images with significant visual distortions or ambiguous content that can be interpreted in multiple ways. CLIP might also misalign images with complex or rare entities.

\textbf{Second Stage: MLLM-based Fine-Grained Filtering.} In the second stage, we employ a Multi-Modal Large Language Model (MLLM) for fine-grained verification. This model performs a deeper analysis of the alignment between the textual entity and the visual features by incorporating contextual information and semantic reasoning. We pass the filtered images from the first stage into the MLLM, where the model evaluates the image-entity alignment in a more refined manner, considering additional contextual clues that CLIP may have missed. 
\subsection{Evaluation}
We conduct all benchmark evaluations using MedEvalKit~\cite{xu2025lingshu}, a unified medical evaluation framework released alongside Lingshu~\cite{xu2025lingshu}. 
MedEvalKit consolidates major multimodal and text-only medical benchmarks into a standardized pipeline, providing consistent dataset adapters, prompt templates, answer parsing, and reporting utilities, thereby enabling fair and reproducible comparisons across different MLLM backbones. 
In our experiments, we follow MedEvalKit's official benchmark suite and configurations, evaluating our fine-tuned model on multimodal medical QA datasets including MMMU-Med~\cite{yue2024mmmu}, VQA-RAD~\cite{lau2018dataset}, SLAKE~\cite{liu2021slake}, PathVQA~\cite{he2020pathvqa}, PMC-VQA~\cite{zhang2023pmc}, and OmniMedVQA~\cite{hu2024omnimedvqa}, with the same unified interface for data loading and metric computation.
For generation-based evaluation, we adopt deterministic decoding (temperature $=0$) as recommended by MedEvalKit to reduce variance across runs; other decoding constraints (e.g., top-$p$ and repetition penalty) follow the toolkit defaults to ensure consistency with prior medical MLLM evaluations.
\section{Synthesized Data Example}
\begin{figure*}[!t]
\centering
\includegraphics[width=1.0\linewidth]{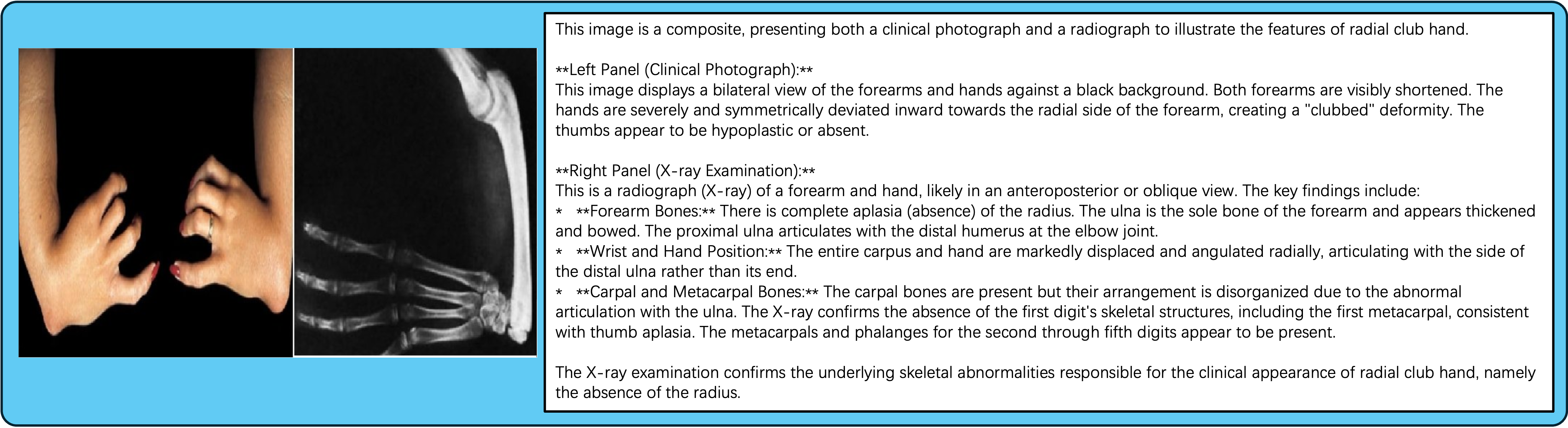}
\caption{An example of the synthesized recaption data.}
\label{supp:fig_data_recaption}
\end{figure*}
\begin{figure*}[!t]
\centering
\includegraphics[width=1.0\linewidth]{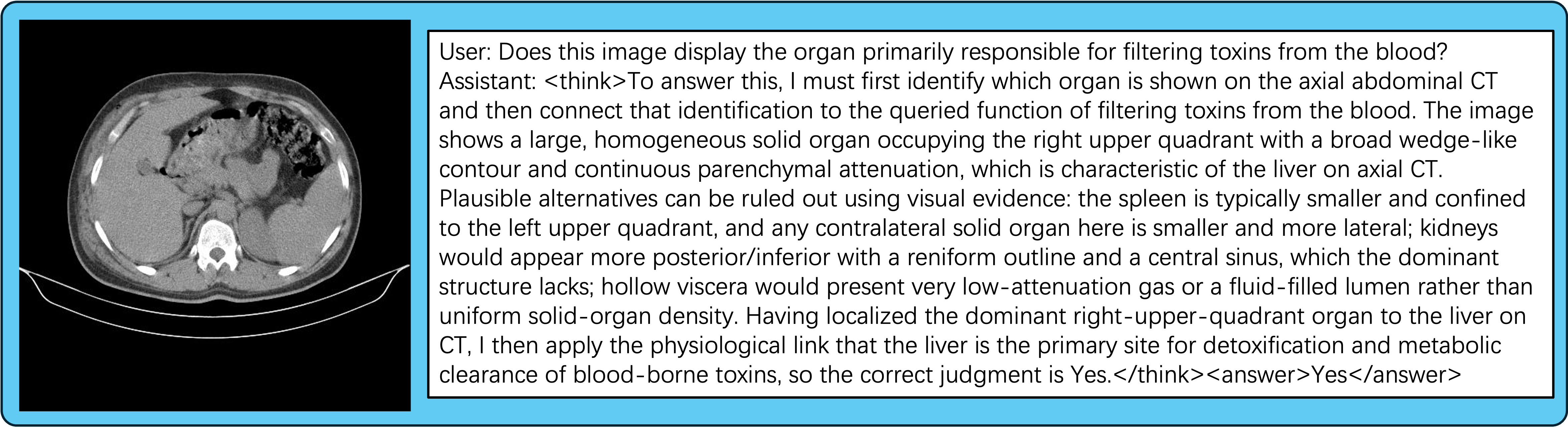}
\caption{An example of the synthesized judgement VQA data.}
\label{supp:fig_data_judgement}
\end{figure*}
\begin{figure*}[!t]
\centering
\includegraphics[width=1.0\linewidth]{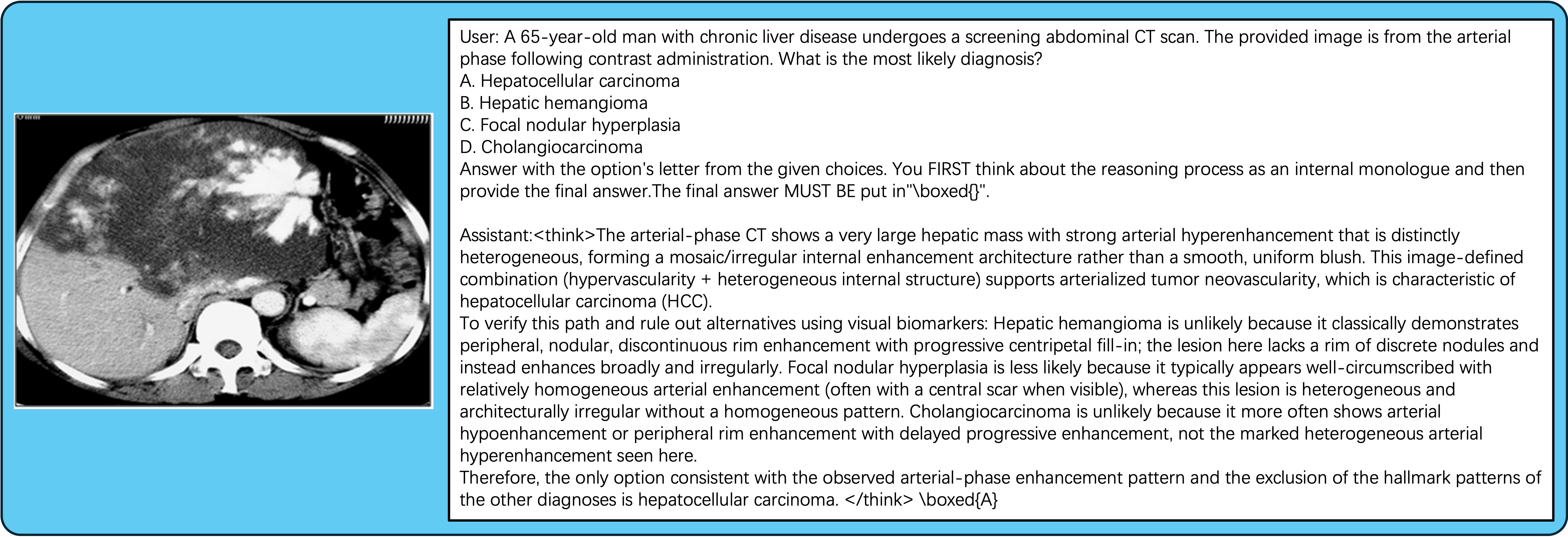}
\caption{An example of the synthesized MCQ VQA data.}
\label{supp:fig_data_mcq}
\vspace{-20pt}
\end{figure*}
Fig.~\ref{supp:fig_data_recaption} shows a representative sample synthesized by our knowledge-aware pipeline for Track~1 (Contextual Re-Captioning), which targets the pervasive issue of sparse or noisy web captions.
The input is a composite figure with a clinical photograph (left) and a corresponding radiograph (right), accompanied by an original caption and a set of MET-linked entities that specify the target condition and its key manifestations (e.g., radial club hand, radial deviation, radius aplasia, thumb hypoplasia).
Conditioned on these hierarchical entities, the generated recaption is strictly grounded in panel-wise visual evidence: it first summarizes the clinical appearance (bilaterally shortened forearms with severe radial deviation and absent/hypoplastic thumbs), and then substantiates these findings with radiographic cues (absence of the radius, abnormal wrist--hand alignment relative to the distal ulna, and missing first-ray skeletal structures).
Beyond listing observations, the recaption explicitly links phenotype to plausible structural etiology by attributing the visible deformity to the underlying skeletal abnormality, thereby injecting precise MET terminology while preserving objective, image-supported phrasing. 
Fig.~\ref{supp:fig_data_judgement} illustrates a reasoning-intensive VQA instance generated by Track~2 (Structure-Constrained Reasoning Synthesis).
Given an abdominal CT image and a MET-derived inference chain, the pipeline formulates a judgement-style question that requires causal reasoning anchored to anatomy recognition rather than text priors.
Here, the model must decide whether the depicted organ is primarily responsible for filtering toxins from the blood.
Using visual biomarkers---a large homogeneous parenchymal organ in the right upper quadrant with a wedge-like contour---the organ is identified as the liver, while alternatives (e.g., spleen or kidneys) are ruled out by their characteristic locations and morphologies on axial CT.
Mapping the recognized anatomical entity to its detoxification role yields the correct judgement, exemplifying how Track~2 couples image-dependent evidence with knowledge-based verification. 
Fig.~\ref{supp:fig_data_mcq} presents an MCQ-style VQA sample produced by the same Track~2 pipeline, where the question asks for the most likely diagnosis among four plausible hepatic lesions (HCC, hemangioma, FNH, cholangiocarcinoma).
The arterial-phase CT demonstrates a large hepatic mass with marked heterogeneous hyperenhancement and an irregular/mosaic internal architecture, supporting arterialized tumor neovascularity characteristic of hepatocellular carcinoma (HCC).
The distractors are excluded by absent hallmark patterns (e.g., no peripheral nodular discontinuous enhancement for hemangioma; no typical homogeneous, well-circumscribed enhancement pattern suggestive of FNH; and no hypoenhancing/rim-enhancing appearance commonly associated with cholangiocarcinoma).
This example highlights that our structure-constrained synthesis enforces strict image dependency and requires explicit visual justification to confirm the target diagnosis while ruling out competing alternatives.
\end{document}